# Novel Force Estimation-based Bilateral Teleoperation applying Type-2 Fuzzy logic and Moving Horizon Estimation

LIAO Qianfang, SUN Da and REN Hongliang

**Abstract** – This paper develops a novel force observer for bilateral teleoperation systems. Type-2 fuzzy logic is used to describe the overall dynamic system, and Moving Horizon Estimation (MHE) is employed to assess clean states as well as the values of dynamic uncertainties, and simultaneously filter out the measurement noises, which guarantee the high degree of accuracy for the observed forces. Compared with the existing methods, the proposed force observer can run without knowing exact mathematical dynamic functions and is robust to different kinds of noises. A force-reflection four-channel teleoperation control laws is also proposed that involving the observed environmental and human force to provide the highly accurate force tracking between the master and the slave in the presence of time delays. Finally, experiments based on two haptic devices demonstrate the superiority of the proposed method through the comparisons with multiple state-to-the-art force observers.

1. Introduction

Bilateral teleoperation systems allow the master and the slave robots to interact with each other through communication channels. It has been found in a wide range of applications such as space exploration, underwater exploration, mining to minimally invasive surgery. Numerous control schemes for teleoperation systems have been proposed to provide stable and relatively accurate position tracking. Such as Proportional Derivative (PD) control methods [21]-[27], sliding mode controllers [28]-[29], robust control strategies [30], adaptive auxiliary switching control methods [31]-[32]. A shortcoming of these methods is that they ignore the force/torque tracking and thus have not considered the transparency. Transparency is the main standard in the bilateral teleoperation control besides stability. High transparency means that the technical medium between master and slave is not felt by the human operator, which not only requires the position of the slave accurately track that of the master, but also requires that the operator can freely drive the master manipulator without feeling any force feedback during free motion, and can receive a perception on the environmental force reflection when the slave contacting an environmental object. If a control method cannot offer the operator an accurate perception about the environment that allows the operator to provide equative force, this control method is not a transparent method and will not achieve satisfactory results in real applications.

It is generally accepted that the four-channel control methods [33]-[34] with force tracking method can achieve higher transparency than other methods. However, the degree of accuracy of the results provided by the force tracking method can seriously influence the stability and transparency of the overall system. In [35]-[37], the human and the environmental forces are assumed to be constant and derived from a known human/environment impedance which does not fit for the reality.

In [38], a force sensor is applied to detect the environmental force. However, force sensors are not always available or feasible in practical applications such as the teleoperated minimally invasive surgery for cardiac procedures. Furthermore, as the external sensors, the force sensors may contain large noises that will cause stability problems especially when transmitting force signals via time-delayed communication channels.

To overcome the above mentioned drawbacks, several studies of "soft sensors" that using different algorithms to estimate the external forces have been proposed. Reaction Force Observer (RFOB) applied in [10]-[14] is a useful tool to estimate the external force in linear robotic system. However, its performance is limited when used in the nonlinear robotic systems since its constant gain cannot cover the overall system dynamics. Also, the applied low-pass filter cannot efficiently suppress white noises. The external disturbance observer proposed in [1]-[9] are designed for nonlinear robotic systems. These force estimation algorithms rely on accurate priori knowledge on the dynamic models and clean measurement of the position and velocity signals. The colorful noises (dynamic uncertainties) and white noises (measurement noises) will largely influence the accuracy of these observers and will further influence the stability of the robotic system if the estimated forces are involved in the system control method. Recently, a novel force estimation method, called Extended Active Observer (EAOB), is applied in [15]-[20] that can efficiently suppress white noises using the extended Kalman filter. [15] also claims that only position signals are required to estimate the force. However, the EAOB also requires an accurate nonlinear dynamic model. Large dynamic uncertainties can have a seriously impact on the velocity estimation and finally make the force estimation fail. In the practical robotic application, ideal dynamic models are impossible to derive to the extent that all of the above force estimation methods have a limited performance. Given this condition, applying these methods to the bilateral teleoperation control can cause the degradation of transparency and even jeopardize the system stability.

In this paper, a novel force observer is proposed to derive accurate external forces by means of two powerful tools: Type-2 fuzzy logic and Moving Horizon Estimation (MHE). The interval Type-2 Takagi-Sugeno (T-S) fuzzy model [45] is used to describe the robotic system. T-S fuzzy model consists of a batch of "If-Then" rules that is using a group of linear local models smoothly blended through fuzzy membership functions to describe a global nonlinear system. This special structure provides the possibility to apply well-developed linear methods on nonlinear systems. Since T-S fuzzy model can be constructed to a high degree of accuracy based on data samples and expert experience [46], the exact mathematical dynamic functions are no longer required. Moreover, fuzzy system is strong in handling disturbances [46]. Compared to the traditional (Type-1) T-S fuzzy model, the interval Type-2 T-S fuzzy model utilizes intervals in lieu of crisp numbers as the fuzzy memberships and local model's coefficients, which endows it additional degree to describe the inexactness and subsequently increased power to deal with uncertainties [45]. On the other hand, MHE is employed to derive the clean signals of position and velocity, as well as the values of dynamic uncertainties, and filter out the measurement noise at the same time. MHE is using optimization-based strategy to estimate states and parameters online based on a finite horizon of the most recent information. At each time step, the information batch is updated by adding the current measures and discarding the oldest ones to move the horizon forward and keep the amount of the stored information fixed. The estimates are derived by minimizing a predefined cost function with respect to the updated information batch. Compared to the famous Kalman filtering methods, MHE works with less computational complexity, and does not necessarily require the noises following Gaussian probability distributions [41]. Thus it can work well in practical operations where the noises are generally diverse and may not met a specified assumption. Reference [41]-[42] demonstrates that the MHE can outperform the Kalman filter in terms of robustness and tracking accuracy. By far, MHE has been applied to different areas using different types of linear and nonlinear discrete-time dynamic functions [39]-[44]. A drawback is that they ignore the fact that linear functions generally struggle to describe a complex system while accurate nonlinear functions are difficult to obtain. In this paper, the usage of fuzzy model provides a solution. To the best of author's knowledge, it is first time to implement MHE based on Type-2 fuzzy model.

By virtue of the information provided by Type-2 fuzzy model and MHE, the force can be estimated through simple calculations. The superiorities of the proposed force observer are: i) Unlike the external disturbance observers proposed in [1]-[9] that requires perfect dynamic models without any disturbances or the RFOB proposed in [10]-[14] only workable for linear systems, the proposed force observer can work for the nonlinear robotic system without requiring any mathematical dynamic functions of the robotic system; ii) Different from the Kalman filter-based force observers in [15]-[20], the proposed force observer can efficiently eliminate both of the colourful noises (system uncertainties) and the white noises (measurement/estimation noises), and does not have the limit of the Kalman filter, poor priori knowledge about the initial points. A Type-2 fuzzy model based four-channel teleoperation control method is given to apply the proposed force observer on bilateral teleoperation system. The experimental results and comparisons with multiple existing force estimating methods demonstrated the accuracy and effectiveness of the proposed force observer.

The rest of the paper is organized as follows: Section 2 describes the Type-2 T-S fuzzy modelling for the system dynamics. Section 3 gives the details of the proposed force observer using Type-2 fuzzy model and MHE. Section 4 introduces the four-channel control laws applying the designed force observer. Section 5 presents the experimental results of the proposed approach and make multiple comparisons with previous work. Conclusions are drawn in Section 6.

Notations: $I_n$ and $0_n$ denote the $n \times n$ identity matrix and $n \times n$ zeros matrix, respectively. $\|X\|^2$ and $\|X\|_P^2$ stand for $X^T X$ and $X^T P X$, respectively, where $P$ is a symmetric positive-semidefinite matrix.

2. Type-2 T-S fuzzy modelling for robotic systems

This section introduces a data-driven approach to identifying an interval Type-2 T-S fuzzy model for a robotic system (master or slave) in bilateral teleoperation. The interval Type-2 T-S fuzzy rules can be expressed as:

Rule $l$: IF $x(k)$ is $\tilde{Z}^l$, THEN

$$\begin{cases} y_1(k) = a_{11}^l x_1(k) + a_{12}^l x_2(k) + \cdots + a_{1q}^l x_q(k) + \tilde{f}_1^l \\ y_2(k) = a_{21}^l x_1(k) + a_{22}^l x_2(k) + \cdots + a_{2q}^l x_q(k) + \tilde{f}_2^l \\ \vdots \\ y_p(k) = a_{p1}^l x_1(k) + a_{p2}^l x_2(k) + \cdots + a_{pq}^l x_q(k) + \tilde{f}_p^l \end{cases} \quad (1)$$

where $l = 1, \cdots, L$, $L$ is the number of fuzzy rules; the premise variables $x_j(k) \in R^n$, $j = 1, \cdots, q$, compose a vector $x(k) = [x_1(k)^T \quad x_2(k)^T \quad \cdots \quad x_q(k)^T]^T \in R^{nq}$; $\tilde{Z}^l$ is an interval Type-2 fuzzy set where the fuzzy membership for $x(k)$ is an interval that can be denoted as $\tilde{\mu}^l(x(k)) = [\underline{\mu}^l(x(k)), \overline{\mu}^l(x(k))]$, $\underline{\mu}^l(x(k))$ and $\overline{\mu}^l(x(k))$ are lower and upper bounds, respectively; $y_i(k) \in R$, $i = 1, \cdots, p$, are outputs and can form a vector $y(k) = [y_1(k) \quad y_2(k) \quad \cdots \quad y_p(k)]^T \in R^p$; $a_{ij}^l = [a_{ij1}^l \quad a_{ij2}^l \quad \cdots \quad a_{ijn}^l] \in R^{1 \times n}$, $i = 1, \cdots, p$, $j = 1, \cdots, q$, are local models' coefficients; $\tilde{f}_i^l$, $i = 1, \cdots, p$ are intervals as $\tilde{f}_i^l = [\underline{f}_i^l, \overline{f}_i^l]$, $\underline{f}_i^l, \overline{f}_i^l \in R$ are lower and upper bounds respectively that indicate the ranges of uncertainties in $y_i(k)$. Blending the $L$ Type-2 fuzzy rules has:

$$\begin{cases} y_1(k) = a_{11}(k)x_1(k) + a_{12}(k)x_2(k) + \cdots + a_{1q}(k)x_q(k) + \tilde{f}_1(k) \\ y_2(k) = a_{21}(k)x_1(k) + a_{22}(k)x_2(k) + \cdots + a_{2q}(k)x_q(k) + \tilde{f}_2(k) \\ \vdots \\ y_p(k) = a_{p1}(k)x_1(k) + a_{p2}(k)x_2(k) + \cdots + a_{pq}(k)x_q(k) + \tilde{f}_p(k) \end{cases} \quad (2)$$

where $a_{ij}(k) = [a_{ij1}(k) \quad a_{ij2}(k) \quad \cdots \quad a_{ijn}(k)] \in R^{1 \times n}$, $a_{ijr}(k) = \frac{1}{2}\left(\frac{\sum_{l=1}^L \underline{\mu}^l(x(k))a_{ijr}^l}{\sum_{l=1}^L \underline{\mu}^l(x(k))} + \frac{\sum_{l=1}^L \overline{\mu}^l(x(k))a_{ijr}^l}{\sum_{l=1}^L \overline{\mu}^l(x(k))}\right)$, $i = 1, \cdots, p$, $j = 1, \cdots, q$ and $r = 1, \cdots, n$; the intervals $\tilde{f}_i(k) = [\underline{f}_i(k), \overline{f}_i(k)]$, where $\underline{f}_i(k) = \frac{\sum_{l=1}^L \underline{\mu}^l(x(k))\underline{f}_i^l}{\sum_{l=1}^L \underline{\mu}^l(x(k))}$, $\overline{f}_i(k) = \frac{\sum_{l=1}^L \overline{\mu}^l(x(k))\overline{f}_i^l}{\sum_{l=1}^L \overline{\mu}^l(x(k))}$, $i = 1, \cdots, p$, can be rewritten as:

$$\tilde{f}_i(k) = f_i(k) + \Delta f_i(k)\lambda_i(k) \quad (3)$$

where $f_i(k) = \frac{\overline{f}_i(k) + \underline{f}_i(k)}{2}$, $\Delta f_i(k) = \frac{\overline{f}_i(k) - \underline{f}_i(k)}{2}$, and $\lambda(k)$ is an unknown parameter satisfying $-1 \leq \lambda(k) \leq 1$.

The interval Type-2 T-S fuzzy model in (1) can be constructed from data samples with assistance of human experience. Denote the input-output data collected from the original system as $z(k) = [x(k)^T \quad y(k)^T]^T \in R^{nq+p}$, $k = 1, \cdots, N_s$, $N_s$ is the number of data, Gustafson-Kessel (G-K) clustering algorithm [47] is employed to identify the fuzzy sets, and weighted least square method is selected to calculate the local model's coefficients. The steps are introduced as follows:

i). Given the number of fuzzy rules as $L$, applying G-K clustering algorithm on the data samples can, firstly, locate $L$ fuzzy cluster centers, denoted by $z_c^l = [(x_c^l)^T \quad (y_c^l)^T]^T$, $l = 1, \cdots, L$, where $x_c^l = [(x_{1c}^l)^T \quad \cdots \quad (x_{qc}^l)^T]^T$, $x_{jc}^l \in R^n$, $j = 1, \cdots, q$, and $y_c^l \in R^p$ are centers for $x(k)$ and $y(k)$ in $l$ th fuzzy cluster, respectively; secondly, give a crisp fuzzy membership to each sample in each fuzzy cluster, denoted by $\mu^l(z(k))$, $l = 1, \cdots, L$, which satisfying $0 \leq \mu^l(z(k)) \leq 1$ and $\sum_{l=1}^L \mu^l(z(k)) = 1$.

ii). For the $l$th fuzzy cluster, determine a blurring radius for the fuzzy memberships, denoted by $\Delta\mu^l$, $0 \leq \Delta\mu^l \leq 1$, to extend each crisp fuzzy membership $\mu^l(z(k))$ to an interval $\tilde{\mu}^l(z(k)) = [\underline{\mu}^l(z(k)), \overline{\mu}^l(z(k))]$ by:

$$\begin{cases} \underline{\mu}^l(z(k)) = \max\{0, \quad \mu^l(z(k)) - \Delta\mu^l\} \\ \overline{\mu}^l(z(k)) = \min\{\mu^l(z(k)) + \Delta\mu^l, \quad 1\} \end{cases} \quad (4)$$

iii). For each fuzzy rule, identify $p$ linear polynomials as in (1) except each $f_i^l$ is replaced by a crisp number $f_i^l$:

$$y_i(k) = a_{i1}^l x_1(k) + a_{i2}^l x_2(k) + \cdots + a_{iq}^l x_q(k) + f_i^l \quad (5)$$

$i = 1, \cdots, p$, through using weighted least square method where the crisp fuzzy memberships are used as weights:

$$a_i^l = (X^T W^l X)^{-1} X^T W^l Y_i \quad (6)$$

where $a_i^l = [a_{i1}^l \cdots a_{iq}^l \quad f_i^l]^T \in R^{nq+1}$, $X = \begin{bmatrix} x_1(1)^T & \cdots & x_q(1)^T & 1 \\ x_1(2)^T & \cdots & x_q(2)^T & 1 \\ \vdots & \vdots & \vdots & \vdots \\ x_1(N_s)^T & \cdots & x_q(N_s)^T & 1 \end{bmatrix} \in R^{N_s \times (nq+1)}$, $Y_i = [y_i(1) \quad y_i(2) \quad \cdots \quad y_i(N_s)]^T \in R^{N_s}$, $W^l = diag[\mu^l(z(1)) \quad \mu^l(z(2)) \quad \cdots \quad \mu^l(z(N_s))]$.

iv). For each fuzzy rule, determine a blurring radius for each output, denoted by $\Delta y_i^l \geq 0$, $i = 1, \cdots, p$, to turn each $f_i^l$ to an interval $\tilde{f}_i^l = [\underline{f}_i^l, \overline{f}_i^l]$ by $\underline{f}_i^l = f_i^l - \Delta y_i^l$ and $\overline{f}_i^l = f_i^l + \Delta y_i^l$. The interval Type-2 T-S fuzzy modeling is complete.

When given a new premise variable vector $x(k)$, the $L$ fuzzy cluster centers are used to calculated the crisp fuzzy memberships:

$$\mu^l(x(k)) = \begin{cases} 0, & \forall D^2(x(k), x_c^v) = 0, v = 1, \ldots, L, v \neq l \\ \frac{1}{\sum_{v=1}^L \frac{D^2(x(k), x_c^l)}{D^2(x(k), x_c^v)}}, & D^2(x(k), x_c^v) \neq 0, for\ v = 1, \ldots, L \\ 1, & D^2(x(k), x_c^l) = 0 \end{cases}$$

$l = 1, \cdots, L \quad (7)$

where $D^2(x(k), x_c^l) = \|x(k) - x_c^l\|^2$ is the Euclidean distance between $x(k)$ and $x_c^l$. The interval Type-2 fuzzy membership for $x(k)$, $\tilde{\mu}^l(x(k)) = [\underline{\mu}^l(x(k)), \overline{\mu}^l(x(k))]$, is derived by:

$$\begin{cases} \underline{\mu}^l(x(k)) = \max\{0, \quad \mu^l(x(k)) - \Delta\mu^l\} \\ \overline{\mu}^l(x(k)) = \min\{\mu^l(x(k)) + \Delta\mu^l, \quad 1\} \end{cases}$$

Afterwards the model's outputs can be obtained by (2) and (3).

**Remark 1**: In the general methods to defuzzify an interval Type-2 fuzzy model, firstly, a type-reduced set, which is an interval of output, is firstly calculated using blending algorithms such as the Karnik–Mendel iterative calculation [45]. Afterwards, the center of the type-reduced set is used as the crisp output. However, for an interval Type-2 fuzzy model, each point in the interval has the same probability to appear. Thus it may not be accurate to use the midpoint as the crisp output. In this paper, by submitting (3) to (2), we can have a more appropriate expression for the output. The unknown coefficient $\lambda(k)$ can be estimated using the MHE presented later.

The dynamics of an $n$-degree of freedom (DOF) robotic system generally can be described by the following equation:

$$M(q)\ddot{q} + C(q, \dot{q})\dot{q} + g(\dot{q}) + f\dot{q} + f_c(\dot{q}) + F^d = \tau + \tau_e \quad (8)$$

where $\ddot{q}, \dot{q}, q \in R^n$ stand for the vectors of joint acceleration, velocity and position signals, respectively. $M(q) \in R^{n \times n}$ and $C(q, \dot{q}) \in R^{n \times n}$ are the inertia matrices and Coriolis/centrifugal effects, respectively. $g(\dot{q}) \in R^n$ is the gravitational force. $f\dot{q} \in R^n$ and $f_c(\dot{q}) \in R^n$ denote the viscous friction and Coulomb friction, respectively. $F^d \in R^n$ contains the unknown disturbance and measurement noise. $\tau = [\tau_1 \quad \cdots \quad \tau_n]^T \in R^n$ are control variables, $\tau_e \in R^n$ is external torque. The following Type-2 T-S fuzzy rules are chosen to describe the $n$-DOF robotic system in free motion:

Rule $l$: IF $x(k)$ is $\tilde{Z}^l$, THEN

$$M^l \frac{v(k+1) - v(k)}{\Delta T} + C^l v(k) + D^l q(k) + \tilde{F}^l = \tau(k) \quad (9)$$

where $l = 1, \cdots, L$, the premise variable vector $x(k) \in R^{3n}$ is $x(k) = [v(k+1)^T \quad v(k)^T \quad q(k)^T]^T$; $q(k) \in R^n$ is the position signal at time $k$ derived from the encoder, and $v(k) \in R^n$ denotes the velocity at time $k$ provided by an observer to be defined later; $\Delta T$ is the sampling time; $M^l, C^l, D^l \in R^{n \times n}$ are coefficients for the local model of $l$th fuzzy rule, $M^l = [M_{ij}^l]_{n \times n}$, $C^l = [C_{ij}^l]_{n \times n}$ and $D^l = [D_{ij}^l]_{n \times n}$. $\tilde{F}^l = [\tilde{F}_i^l]_{n \times 1}$, $\tilde{F}_i^l = [\underline{F}_i^l, \overline{F}_i^l]$, $\underline{F}_i^l$ and $\overline{F}_i^l$ are lower and upper bounds respectively. The $l$th fuzzy cluster center is denoted by $z_c^l = [(x_c^l)^T \quad (\tau_c^l)^T]^T \in R^{4n}$, where $x_c^l = [(v\_1_c^l)^T \quad (v_c^l)^T \quad (q_c^l)^T]^T \in R^{3n}$, and $v\_1_c^l, v_c^l, q_c^l$ and $\tau_c^l$ are centers for $v(k+1), v(k), q(k)$ and $\tau(k)$, respectively. Blending the $L$ fuzzy rules has:

$$M(k)\frac{v(k+1) - v(k)}{\Delta T} + C(k)v(k) + D(k)q(k) + F(k) + \Delta F(k)\lambda(k) = \tau(k) \quad (10)$$

where $M(k) = [M_{ij}(k)]_{n \times n} = \frac{1}{2}\left(\frac{\sum_{l=1}^L \underline{\mu}^l(x(k))M^l}{\sum_{l=1}^L \underline{\mu}^l(x(k))} + \frac{\sum_{l=1}^L \overline{\mu}^l(x(k))M^l}{\sum_{l=1}^L \overline{\mu}^l(x(k))}\right)$, $C(k) = [C_{ij}(k)]_{n \times n} = \frac{1}{2}\left(\frac{\sum_{l=1}^L \underline{\mu}^l(x(k))C^l}{\sum_{l=1}^L \underline{\mu}^l(x(k))} + \frac{\sum_{l=1}^L \overline{\mu}^l(x(k))C^l}{\sum_{l=1}^L \overline{\mu}^l(x(k))}\right)$, $D(k) = [D_{ij}(k)]_{n \times n} = \frac{1}{2}\left(\frac{\sum_{l=1}^L \underline{\mu}^l(x(k))D^l}{\sum_{l=1}^L \underline{\mu}^l(x(k))} + \frac{\sum_{l=1}^L \overline{\mu}^l(x(k))D^l}{\sum_{l=1}^L \overline{\mu}^l(x(k))}\right)$, $F(k) = [F_i(k)]_{n \times 1} = \frac{1}{2}\left(\frac{\sum_{l=1}^L \underline{\mu}^l(x(k))\underline{F}_i^l}{\sum_{l=1}^L \underline{\mu}^l(x(k))} + \frac{\sum_{l=1}^L \overline{\mu}^l(x(k))\overline{F}_i^l}{\sum_{l=1}^L \overline{\mu}^l(x(k))}\right)$, $\Delta F(k) = diag\{\Delta F_1(k), \cdots, \Delta F_n(k)\}$, $\Delta F_i(k) = \frac{1}{2}\left(\frac{\sum_{l=1}^L \overline{\mu}^l(x(k))\overline{F}_i^l}{\sum_{l=1}^L \overline{\mu}^l(x(k))} - \frac{\sum_{l=1}^L \underline{\mu}^l(x(k))\underline{F}_i^l}{\sum_{l=1}^L \underline{\mu}^l(x(k))}\right)$, $\lambda(k) = [\lambda_1(k) \quad \cdots \quad \lambda_n(k)]^T \in R^n$, and $-1 \leq \lambda_i(k) \leq 1$, $i = 1, \cdots, n$. The term $\Delta F(k)\lambda(k)$ in (10) denotes the dynamic uncertainties (colourful noise) that vary with lower and upper bounds. Note that $v(k)$ is a measured velocity signal which actually is the real velocity $\dot{q}$ with measurement noises (white noise). Since $\Delta T$ for a robotic system is generally sufficiently small, $\frac{v(k+1)-v(k)}{\Delta T}$ is able to represent the acceleration $\ddot{q}$ such that (10) can approximate the following continuous function:

$$M\ddot{q} + C\dot{q} + Dq + F + \Delta F\lambda + \eta = \tau \quad (11)$$

where $\eta$ is the measurement noises. Note that (11) will be used to observe the external force/torque $\tau_e$ in the following section.

Based on (10), we have the following discrete-time function to describe the $n$-DOF robotic system:

$$X(k+1) = \mathcal{A}_k X(k) + \mathcal{B}_k u(k) + \mathcal{F}_k \lambda(k)$$

$$y(k) = \mathcal{C}_k X(k) + \eta_k \quad (12)$$

where $X(k) = [q(k)^T \ v(k)^T]^T \in R^{2n}$ is the state vector, $u(k) = \tau(k) - F(k) \in R^n$ is the input; the term $\mathcal{F}_k \lambda(k)$ describes the uncertainty including coloured and white noise, where $\lambda(k) = [\lambda_1(k) \ \cdots \ \lambda_n(k)]^T$; $y(k) = [q(k)^T \ v(k)^T]^T \in R^{2n}$ is the output; the measurement noises $\eta_k \in R^{2n}$. $\mathcal{A}_k$, $\mathcal{B}_k$, $\mathcal{C}_k$ and $\mathcal{F}_k$ are time-varying coefficient matrices with proper dimensions as

$$\mathcal{A}_k = \begin{bmatrix} I_n & \Delta T I_n \\ -\Delta T M^{-1}(k)D(k) & I_n - \Delta T M^{-1}(k)C(k) \end{bmatrix}, \quad \mathcal{B}_k = \begin{bmatrix} 0_n \\ \Delta T M^{-1}(k) \end{bmatrix}, \quad \mathcal{C}_k = \begin{bmatrix} I_n & 0_n \\ 0_n & I_n \end{bmatrix}, \text{ and } \mathcal{F}_k = \begin{bmatrix} 0_n \\ -\Delta T M^{-1}(k)\Delta F(k) \end{bmatrix}.$$

3. Force observer based on Type-2 fuzzy logic and moving horizon estimation

This section introduces the force observer for the bilateral teleoperation systems that is able to derive the environmental force/torque estimation, denoted by $\hat{\tau}_e$, from 1). the joint position signals, 2). the joint velocity signals, and 3). the information in (11). The position signals are provided by the position encoders, and the velocity signals can be obtained using the following velocity observer:

$$\dot{v} = k_{v1}(q - v) + k_{v2}(q - v) + k_{v1}k_{v2}\int_0^t (q(\iota) - v(\iota))d\iota \quad (13)$$

where $k_{v1}$ and $k_{v2}$ are the constants. The estimated velocity signals have two usages: i) they are used to compose the data samples to identify the Type-2 T-S fuzzy model; ii) they are supervising signals for MHE which will be introduced later. Note that the estimated velocity signals will not be involved in the control law designs.

*Theorem 1*: The velocity observed by (13) can track its real value $\dot{q}$. The convergence error $\Delta e_v = q - v$ is bounded and is neighbouring zero at the steady state.

*Proof*: See Appendix.

The information in (11) can be obtained from the identified Type-2 T-S fuzzy model, except the term $\Delta F \lambda$ to describe dynamic uncertainty is unknown. In addition, the measured position signals and the estimated velocity signals from (13) contain noises, which prevent the force observer providing accurate results. In order to guarantee a good performance of the force observer, we introduce a Type-2 T-S fuzzy model based MHE method to derive the clean signals of position and velocity, and determine the value of $\Delta F \lambda$ in (11).

MHE is an effective method that using optimization techniques to estimate system states and parameter based on a fixed horizon of system measures with noises. The Type-2 fuzzy model in (12), where the state $X(k) = [q(k)^T \ v(k)^T]^T$ and the parameter $\lambda(k)$ are exactly the information needed for force observer, plays an important role in the MHE implementation. Given the horizon size as $N$, a batch of outputs and inputs defined by (12) can be observed from the original $n$-DOF system are available at any time $k > N$ that denoted by $\mathcal{I}_k = \{y_k^{k-N}, u_{k-1}^{k-N}\}$, where

$$y_k^{k-N} = [y(k-N)^T \ y(k-N+1)^T \ \cdots \ y(k)^T]^T$$
$$u_{k-1}^{k-N} = [u(k-N)^T \ u(k-N+1)^T \ \cdots \ u(k-1)^T]^T$$

Suppose the pair $(\mathcal{C}_k, \mathcal{A}_k)$ in (12) is completely observable in $N$ steps, the estimated states and parameters at time $k$ are expressed by:

$$\hat{X}_{k|k}^{k-N|k} = [\hat{X}(k-N|k)^T \ \cdots \ \hat{X}(k-1|k)^T]^T$$
$$\hat{\lambda}_{k-1|k}^{k-N|k} = [\hat{\lambda}(k-N|k)^T \ \cdots \ \hat{\lambda}(k-1|k)^T]^T$$

The following notations will be used in the remainder of this paper:

$$\lambda_{k-1}^{k-N} = [\lambda(k-N) \ \lambda(k-N+1) \ \cdots \ \lambda(k-1)]^T,$$
$$\eta_k^{k-N} = [\eta(k-N) \ \eta(k-N+1) \ \cdots \ \eta(k)]^T,$$

$$\Phi_k = \begin{bmatrix} \mathcal{C}_{k-N} \\ \mathcal{C}_{k-N+1}\mathcal{A}_{k-N} \\ \mathcal{C}_{k-N+2}\mathcal{A}_{k-N+1}\mathcal{A}_{k-N} \\ \vdots \\ \mathcal{C}_k\mathcal{A}_{k-1}\mathcal{A}_{k-2}\cdots\mathcal{A}_{k-N} \end{bmatrix},$$

$$G_k = \begin{bmatrix} 0 & 0 & \cdots & 0 \\ \mathcal{C}_{k-N+1}\mathcal{B}_{k-N} & 0 & \cdots & 0 \\ \mathcal{C}_{k-N+2}\mathcal{A}_{k-N+1}\mathcal{B}_{k-N} & \mathcal{C}_{k-N+2}\mathcal{B}_{k-N+1} & \cdots & 0 \\ \vdots & \vdots & \vdots & \vdots \\ g_{N+1,1} & g_{N+1,2} & \cdots & \mathcal{C}_k\mathcal{B}_{k-1} \end{bmatrix},$$

$$\mathcal{H}_k = \begin{bmatrix} 0 & 0 & \cdots & 0 \\ \mathcal{C}_{k-N+1}\mathcal{F}_{k-N} & 0 & \cdots & 0 \\ \mathcal{C}_{k-N+2}\mathcal{A}_{k-N+1}\mathcal{F}_{k-N} & \mathcal{C}_{k-N+2}\mathcal{F}_{k-N+1} & \cdots & 0 \\ \vdots & \vdots & \vdots & \vdots \\ h_{N+1,1} & h_{N+1,2} & \cdots & \mathcal{C}_k\mathcal{F}_{k-1} \end{bmatrix},$$

where $g_{N+1,1} = \mathcal{C}_k\mathcal{A}_{k-1}\mathcal{A}_{k-2}\cdots\mathcal{A}_{k-N+1}\mathcal{B}_{k-N}$,
$g_{N+1,2} = \mathcal{C}_k\mathcal{A}_{k-1}\mathcal{A}_{k-2}\cdots\mathcal{A}_{k-N+2}\mathcal{B}_{k-N+1}$,
$h_{N+1,1} = \mathcal{C}_k\mathcal{A}_{k-1}\mathcal{A}_{k-2}\cdots\mathcal{A}_{k-N+1}\mathcal{F}_{k-N}$,
$h_{N+1,2} = \mathcal{C}_k\mathcal{A}_{k-1}\mathcal{A}_{k-2}\cdots\mathcal{A}_{k-N+2}\mathcal{F}_{k-N+1}$.

The MHE optimization problem is stated as follows:

*Problem 1*: At any time $k > N$, given the information $\{\mathcal{I}_k, \overline{X}(k-N)\}$, where $\overline{X}(k-N)$ is a priori prediction of state $X(k-N)$ and is generally determined by the estimated value at time $k-1$:

$$\overline{X}(k-N) = \hat{X}(k-N|k-1) \quad (14)$$

find the optimal estimates of $\hat{X}(k-N|k)$ and $\hat{\lambda}_{k-1|k}^{k-N|k}$ by minimizing the following cost function:

$$J(k) = \|\hat{X}(k-N|k) - \overline{X}(k-N)\|_{P_X}^2 + \sum_{i=k-N}^{k-1}\|\lambda(i|k)\|_{P_\lambda}^2 + \sum_{i=k-N}^{k}\|y(i) - \hat{y}(i|k)\|_{P_y}^2 \quad (15)$$

with

$$\hat{X}(i+1|k) = \mathcal{A}_i\hat{X}(i|k) + \mathcal{B}_i u(i) + \mathcal{F}_i\lambda(i|k)$$
$$\hat{y}(i|k) = \mathcal{C}_i\hat{X}(i|k) \quad (16)$$

where $P_X \in R^{2n\times 2n}$, $P_\lambda \in R^{n\times n}$, $P_y \in R^{2n\times 2n}$ are the weight matrices which are positive semi-definite and symmetric. $\hat{y}(i|k)$, $i = k-N, \cdots, k$, are the estimated outputs.

The optimization of Problem 1 gives the following theorem:

*Theorem 2*: when the weight matrix $P_X$, $P_\lambda$ and $P_y$ are given, the solutions to Problem 1 are:

$$\hat{X}(k-N|k) = (P_X + \Phi_k^T \overline{P}_y \Phi_k - \Theta_k \Phi_k)^{-1}\{[\Phi_k^T \overline{P}_y -$$

$\Theta_k](y_k^{k-N} - G_k u_{k-1}^{k-N}) + P_X \overline{X}(k - N)\}$ (17)

$\hat{\lambda}_{k-1|k}^{k-N|k} = \Gamma_k[(y_k^{k-N} - G_k u_{k-1}^{k-N}) - \Phi_k \hat{X}(k-N|k)]$ (18)

where $\Gamma_k = (\overline{P}_\lambda + \mathcal{H}_k^T \overline{P}_y \mathcal{H}_k)^{-1} \mathcal{H}_k^T \overline{P}_y$, $\Theta_k = \Phi_k^T \overline{P}_y \mathcal{H}_k \Gamma_k$, $\overline{P}_\lambda = I_N \otimes P_\lambda$ and $\overline{P}_y = I_{N+1} \otimes P_y$, $\otimes$ is the Kronecker product.

**Proof**: by defining

$\hat{y}_{k|k}^{k-N|k} = [\hat{y}(k-N|k)^T \quad \hat{y}(k-N+1|k)^T \quad \cdots \quad \hat{y}(k|k)^T]^T$,

the cost function can be rewritten as

$J(k) = \|\hat{X}(k-N|k) - \overline{X}(k-N)\|_{P_X}^2 + \|\hat{\lambda}_{k-1|k}^{k-N|k}\|_{\overline{P}_\lambda}^2 + \|y_k^{k-N} - \hat{y}_{k|k}^{k-N|k}\|_{\overline{P}_y}^2$ (19)

where $\hat{y}_{k|k}^{k-N|k}$ can be expressed by the following equation according to (16):

$\hat{y}_{k|k}^{k-N|k} = \Phi_k \hat{X}(k-N|k) + G_k u_{k-1}^{k-N} + \mathcal{H}_k \hat{\lambda}_{k-1|k}^{k-N|k}$ (20)

The following two equations should be satisfied to minimize the cost function:

$\frac{\partial J(k)}{\partial \hat{x}(k-N|k)} = 0, \quad \frac{\partial J(k)}{\partial \hat{\lambda}_{k-1|k}^{k-N|k}} = 0$ (21)

From (19) and (20), the two partial differential functions in (21) can be expressed by:

$\frac{\partial J(k)}{\partial \hat{x}(k-N|k)} = 2P_X(\hat{X}(k-N|k) - \overline{X}(k-N)) - 2\Phi_k^T \overline{P}_y(y_k^{k-N} - \Phi_k \hat{X}(k-N|k) - G_k u_{k-1}^{k-N} - \mathcal{H}_k \hat{\lambda}_{k-1|k}^{k-N|k})$ (22)

$\frac{\partial J(k)}{\partial \hat{\lambda}_{k-1}^{k-N}} = 2\overline{P}_\lambda(\hat{\lambda}_{k-1|k}^{k-N|k}) - 2\mathcal{H}_k^T \overline{P}_y(y_k^{k-N} - \Phi_k \hat{X}(k-N|k) - G_k u_{k-1}^{k-N} - \mathcal{H}_k \hat{\lambda}_{k-1|k}^{k-N|k})$ (23)

Considering (21)-(23) gives:

$P_X \hat{X}(k-N|k) - P_X \overline{X}(k-N) - \Phi_k^T \overline{P}_y(y_k^{k-N} - G_k u_{k-1}^{k-N}) + \Phi_k^T \overline{P}_y \Phi_k \hat{X}(k-N|k) + \Phi_k^T \overline{P}_y \mathcal{H}_k \hat{\lambda}_{k-1|k}^{k-N|k} = 0$ (24)

$\overline{P}_\lambda \hat{\lambda}_{k-1|k}^{k-N|k} - \mathcal{H}_k^T \overline{P}_y(y_k^{k-N} - G_k u_{k-1}^{k-N}) + \mathcal{H}_k^T \overline{P}_y \Phi_k \hat{X}(k-N|k) + \mathcal{H}_k^T \overline{P}_y \mathcal{H}_k \hat{\lambda}_{k-1|k}^{k-N|k} = 0$ (25)

From (24) and (25), it is easy have the solutions of $\hat{X}(k-N|k)$ and $\hat{\lambda}_{k-1|k}^{k-N|k}$ in (17) and (18). Q.E.D.

The rest of the estimated states in $\hat{X}_{k|k}^{k-N|k}$ can be calculated by submitting $\hat{X}(k-N|k)$ and $\hat{\lambda}_{k-1|k}^{k-N|k}$ to (16). The convergence of the state estimates using MHE can be analysed using *Input-to-State Stability* [48] given as follows:

Definition 1 [48]: a nonlinear discrete-time system with external disturbance:

$e(k+1) = f(e(k), \xi(k)), e(k) \in R^{n_e}, \xi(k) \in R^{n_\xi}$ (26)

is *input-to-stable* if and only if it admits a continuous Lyapunov function $V_{e_X}: R^{n_e} \to R_{\geq 0}$ that for $\mathcal{K}_\infty$-functions $\alpha_1, \alpha_2, \alpha_3$ and a $\mathcal{K}$-functions $\sigma$ the following holds

i). $\alpha_1(\|e(k)\|) \leq V_{e_X}(e(k)) \leq \alpha_2(\|e(k)\|)$

ii). $V_{e_X}(e(k)) - V_{e_X}(e(k-1)) \leq -\alpha_3(\|e(k-1)\|) +$

$\sigma(\|\xi(k)\|)$

for all $e(k) \in R^{n_e}$ and $\xi(k) \in R^{n_\xi}$.

Define the error of the estimated states from MHE as:

$e_X(k-N) = X(k-N) - \hat{X}(k-N|k)$ (27)

Submitting (12) and (17) into (27) can have

$e_X(k-N) = X(k-N) - \hat{X}(k-N|k)$
$= (P_X + \Phi_k^T \overline{P}_y \Phi_k - \Theta_k \Phi_k)^{-1} \{P_X(X(k-N) - \overline{X}(k-N)) - (\Phi_k^T \overline{P}_y - \Theta_k)(\mathcal{H}_k \lambda_{k-1}^{k-N} + \eta_k^{k-N})\}$ (28)

According to (12), (14) and (16),

$X(k-N) - \overline{X}(k-N) = X(k-N) - \hat{X}(k-N|k-1)$
$= \mathcal{A}_{k-N-1} e_X(k-N-1) + \mathcal{F}_{k-N-1}[\lambda(k-N-1) - \hat{\lambda}(k-N-1|k-1)]$

then (28) becomes:

$e_X(k-N) = \Psi_k e_X(k-N-1) + \xi_k$ (29)

where $\Psi_k = (P_X + \Phi_k^T \overline{P}_y \Phi_k - \Theta_k \Phi_k)^{-1} P_X \mathcal{A}_{k-N-1}$, $\xi_k = (P_X + \Phi_k^T \overline{P}_y \Phi_k - \Theta_k \Phi_k)^{-1} P_X \mathcal{F}_{k-N-1}[\lambda(k-N-1) - \hat{\lambda}(k-N-1|k-1)] - (P_X + \Phi_k^T \overline{P}_y \Phi_k - \Theta_k \Phi_k)^{-1} (\Phi_k^T \overline{P}_y - \Theta_k)(\mathcal{H}_k \lambda_{k-1}^{k-N} + \eta_k^{k-N})$.

*Theorem 3*: For the system (12) with estimate errors as in (29), if proper $P_X$, $P_\lambda$ and $P_y$ exist and satisfy

$$\Psi_k^T(P_X + I_{2n})\Psi_k - P_X \leq -Q_X$$

where $Q_X$ is a positive semi-definite and symmetric matrix. Then $e_X(k-N)$ is *input-to-state stable*.

*Proof*: see Appendix.

Based on the clean position and velocity signals, and the value of $\Delta F \lambda$ in (11) provided from MHE, the force observer to derive the environmental force/torque estimation $\tau_e^*$ can be introduced:

$\tau_e^* = Z_e + \mathcal{B}_e$

$\mathcal{B}_e = \sigma \hat{q}, \dot{\mathcal{B}}_e = Y_e M \hat{q}$

$Y_e = M^{-1} \sigma$ (30)

$\dot{Z}_e = -Y_e Z_e - Y_e(\tau - C\hat{q} - D\hat{q} - F - \Delta F \lambda + \mathcal{B}_e) + \aleph \hat{q}$ (31)

where $\hat{\dot{q}}$ and $\hat{q}$ are velocity and position estimated by MHE. $\sigma$ is a positive constant. $\aleph$ is a constant gain. Setting different values of $\aleph$ can influence $\tau_e^*$. Therefore, we denote $\tau_{e1}^*$, $\tau_{e2}^*$, ..., $\tau_{eN}^*$ that are determined by $\aleph_{e1}\hat{q}$, $\aleph_{e2}\hat{q}$, ..., $\aleph_{eN}\hat{q}$.

Based on (30), (31) and the dynamic model in (11), and under the condition than the measurement noise $\eta$ is eliminated and the dynamic uncertainty (colourful noise) $\Delta F \lambda$ is known, we can derive the differential of the estimated torque to be

$\dot{\tau}_{eN}^* = \dot{Z}_{eN} + \dot{\mathcal{B}}_e$
$= -Y_e Z_{eN} - Y_e(\tau - C\hat{q} - D\hat{q} - F - \Delta F \lambda + \mathcal{B}_e) + \aleph_{eN}\hat{q} + \dot{\mathcal{B}}_e$
$= Y_e(\tau_e - \tau_{eN}^*) + \aleph_{eN}\hat{q}$ (32)

From (31), when the estimated torque tracks the environmental torque ($\tau_{eN}^* \to \tau_e$), $\dot{\tau}_{eN}^*$ becomes zero at the steady state and $\tau_{eN}^*$ maintain constant. Otherwise, $\dot{\tau}_{eN}^*$ is not zero that allows $\tau_{eN}^*$ keeps increasing till the estimated torque $\tau_{eN}^*$ can track $\tau_e$.

Thus, the differential of the torque estimation errors $\Delta \tau_{eN}$ that $\Delta \tau_{eN} = \tau_e - \tau_{eN}^*$ can be derived as:

$$\Delta \dot{\tau}_{eN} = \dot{\tau}_e - \dot{\tau}_{eN}^* = \dot{\tau}_e - \dot{Z}_{eN} - \dot{\mathcal{B}}_e = \dot{\tau}_e - \Upsilon_e \Delta \tau_{eN} - \aleph_{eN} \hat{\dot{q}} \quad (33)$$

**Remark 2.** In the whole procedure of force observation, only the position signals come from hard sensors (encoder) while others are derived by different estimating algorithms, which greatly save the cost and space to install hardware sensor on the robotic system.

The proposed Type-2 fuzzy logic and MHE based force observer eliminates measurement noises to give clean signals of positions, $\hat{q}$ and velocities, $\hat{\dot{q}}$, as well as the values of dynamic uncertainties $\Delta F \lambda$, and most important, it gives the external force/torque estimates, $\tau_e$. Therefore, the Type-2 fuzzy-based dynamic model in (11) can now be improved and expressed by:

$$M \hat{\ddot{q}} + C \hat{\dot{q}} + D \hat{q} + F + \Delta F \lambda = \tau + \hat{\tau}_e \quad (34)$$

Notice that $\hat{\ddot{q}}$, the differential of $\hat{\dot{q}}$, is not used in further control method design. Therefore, it will not increase the estimation errors.

4. Force reflection bilateral teleoperation

In this section, the proposed force estimation algorithm is applied to the bilateral teleoperation. (34) is expressed in the following teleoperation dynamics (35) by adding the suffix $i = m, s$ which denotes master and slave, and $j = h, e$ denotes human and environmental torques.

$$M_i \hat{\ddot{q}}_i + C_i \hat{\dot{q}}_i + D_i \hat{q}_i + F_i + \Delta F_i \lambda_i = \tau_i + \hat{\tau}_j \quad (35)$$

**Remark 4.** In this paper, the feedforward and feedback time-varying delays $T_1(t)$ and $T_2(t)$ have upper bounds $T_1^{max}$ and $T_2^{max}$, respectively. Also, suppose the differential of the real human and environmental torques $\dot{\tau}_h$ and $\dot{\tau}_e$ have their constant upper bounds $\bar{\xi}_m$ and $\bar{\xi}_s$, respectively.

Combined with the designed force observer, we use the following 4-channel control laws (36)-(37) to evaluate the force reflection performance of the designed force observer.

$$\tau_m(t) = K_m \left( \hat{q}_s(t - T_2(t)) - \hat{q}_m(t) \right) - B_m \hat{\dot{q}}_m(t) + K_h \left( \tau_{h1}^*(t) - \tau_{e1}^*(t - T_2(t)) \right) - C_m \hat{\dot{q}}_m(t) - D_m \hat{q}_m(t) - F_m - \Delta F_m \lambda_m(t) - \tau_{h2}^*(t) \quad (36)$$

$$\tau_s(t) = K_s \left( \hat{q}_m(t - T_1(t)) - \hat{q}_s(t) \right) - B_s \hat{\dot{q}}_s(t) + K_e \left( \tau_{h3}^*(t - T_1(t)) - \tau_{e3}^*(t) \right) - C_s \hat{\dot{q}}_s(t) - D_s \hat{q}_s - F_s - \Delta F_s \lambda_s(t) - \tau_{e2}^*(t) \quad (37)$$

where $K_m$, $K_s$, $B_m$, $B_s$, $K_h$, and $K_e$ are constant control gains. *Theorem 4*: the proposed bilateral teleoperation control system is stable and the master-slave position and torque tracking errors converge to zero at the steady state if the following condition is satisfied.

$$B_m \geq (T_1^{max} + T_2^{max})I, \quad B_s \geq (T_1^{max} + T_2^{max})I, \quad \aleph_{h2} = I, \aleph_{e2} = I, \aleph_{h1} = -K_h, \aleph_{e3} = -\frac{K_m K_e}{K_s}, \aleph_{h3} = 0, \aleph_{e1} = 0 \quad (38)$$

*Proof*: See Appendix.

5. Experimental results

In this section, a series of experimental results are provided. The applied experimental platform consisting of two 3-DOF haptic devices as shown in Fig. 1 validates the proposed bilateral control algorithm in (36)-(37) applying the designed Type-2 fuzzy logic based force observer. All the experiments involve the three joints of the master and the slave robots. However, considering the image size scaling and for readers' convenience, only the results of Joint 2 are given because that the joint 2 of each robot is largely influenced by gravitational force. Using Joint 2 can test the proposed algorithm's capacity in compensating for dynamic uncertainties including gravity and make an apparent comparison with previous algorithms.

In order to demonstrate the superiority of the proposed force observer, we compare the proposed force estimation algorithm to multiple state-of-the-art force observers in previous work (RFOB in [9]-[13], EAOB in [14]-[19], and the nonlinear disturbance observer (NDOB) in [7]-[8]) by applying all of the force observers into the bilateral algorithm in (36)-(37). By using the Type-2 fuzzy logic strategy, the proposed force observer does not require any priori knowledge of dynamic models. However, the EAOB and NDOB in previous work require relatively accurate dynamic models to support their algorithms. Therefore, the dynamic models estimation including Mass, Centrifugal effects and Gravity provided in [9] are applied to support EAOB and NDOB.

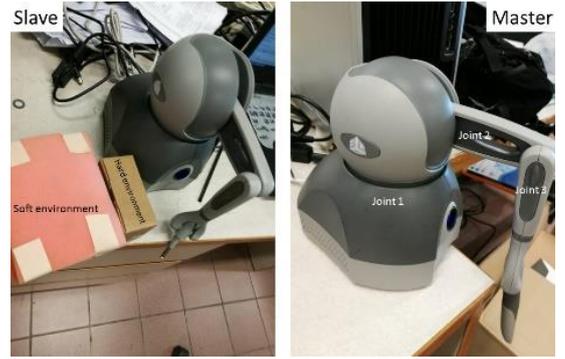

Fig. 1. Experimental platform

The experiment is performed in the presence of different scenarios including free motion, hard contact and soft contact in order to evaluate the performances of the proposed force observer and force observers for comparison. The time delay in the experiment is around 100 ms (one-way) with 50 ms variation. For a Master or a Slave of the teleoperation system with the degree of freedom $n^f = 3$, a Type-2 T-S fuzzy model with $c = 9$ fuzzy rules is constructed from 30409 data samples where the sampling period is $\Delta T = 0.001$. We present the coefficients of the first fuzzy rules of the slave and the master as an example:

For Slave: $x_{s,c}^1 = [-0.2558 \ -0.1726 \ -0.0105 \ 0.1223 \ -0.3325 \ -0.0749 \ -0.2838 \ 0.9580 \ 0.9646]^T$, $\Delta \mu_s^1 = 0.05$,

$$M_s^1 = \begin{bmatrix} 0.2455 & -0.1584 & -0.2090 \\ -0.0044 & 0.2174 & -0.1044 \\ -0.0142 & -0.0057 & 0.0299 \end{bmatrix}, \quad C_s^1 = \begin{bmatrix} -0.0751 & 0.1714 & 0.2668 \\ -0.0180 & -0.1022 & 0.1497 \\ 0.0347 & -0.0210 & 0.0807 \end{bmatrix}, \quad D_s^1 = \begin{bmatrix} 0.1766 & 0.0210 & -0.0739 \\ -0.0439 & 0.0772 & -0.0778 \\ -0.0069 & 0.0175 & -0.0718 \end{bmatrix}, \quad E_s^1 = [-0.1994, -0.2196, -0.1658]^T, \quad \bar{E}_s^1 = [0.3006, 0.2804, 0.3342]^T.$$

For Master: $x_{s,c}^1 = [-0.1724\ 0.0226\ 0.1135\ 0.0726\ -0.5021\ -0.1842\ -0.1030\ 0.9771\ 0.9384]^T$, $\Delta \mu_s^1 = 0.03$,
$M_m^1 = \begin{bmatrix} 0.4188 & -0.0777 & -0.1592 \\ -0.4306 & 0.3593 & -0.1447 \\ -0.0378 & -0.0725 & 0.5974 \end{bmatrix}$, $C_m^1 = \begin{bmatrix} -0.1223 & 0.1532 & 0.0291 \\ -0.1220 & 0.5856 & -0.4906 \\ 0.0736 & -0.0333 & 0.3909 \end{bmatrix}$, $D_m^1 = \begin{bmatrix} 0.0354 & 0.0302 & -0.0044 \\ 0.0057 & 0.0430 & -0.1073 \\ -0.0201 & 0.0202 & 0.0056 \end{bmatrix}$, $\underline{E}_m^1 = [-0.2824, -0.2082, -0.2832]^T$, $\overline{E}_m^1 = [0.3176, 0.3918, 0.3168]^T$.

In the MHE algorithm, $N = 10$ for each robot and the weight matrices are set as $P_X = diag[15,15,15,40,40,40]$, $P_y = diag[30,30,30,100,100,100]$, $P_\lambda = diag[0.35,0.3,0.4,0.3,0.35,0.45]$. According to (38), for the bilateral control algorithm, the parameters are set as $B_m = B_s = diag[0.2,0.2,0.2]$, $K_m = diag[0.1,0.1,0.1]$, $K_s = diag[1,1,1]$, $K_h = diag[0.2,0.2,0.2]$, $K_e = diag[3,3,3]$. $\aleph_{h1} = diag[-0.2,-0.2,-0.2]$, $\aleph_{e3} = diag[-0.3,-0.3,-0.3]$.

5.1 free motion + hard contact

In this sub-section, the master firstly drives the slave to conduct free motion for three rounds and then the slave is controlled to contact to a solid wall. During free motion, an accurate force observer requires the estimated force/torque to be neighbouring zero during free motion since no environmental object is contacted. During hard contact, the output of the observer must fast jump to a certain value to provide the operator a force feedback.

A. RFOB

RFOB is a linear force observer using low-pass filter that does not closely rely on the accurate dynamic model. The key point of RFOB is the configuration of the bandwidth $g_{RFOB}$ of the low-pass filter. Setting $g_{RFOB}$ too large will lose the capacity of noise suppression while setting it too small can degrade the amplitude of the estimated force signal. The value of $g_{RFOB}$ is set to be 500 rad/s as recommended in [10]. The proposed MHE algorithm is used in order to eliminate the noises from the velocity observer. Fig.2 shows the position and torque tracking performance of the system using RFOB. The slave conducts free motion ($0^{th}$ s – $18^{th}$ s) and hard contact ($19^{th}$ s-$23^{rd}$ s). From Fig.2, we can see that RFOB contains several drawbacks. Firstly, the output from the RFOB produces large noises that will significantly affect the perception of the human operator even the noise from the velocity estimator has already been eliminated by the proposed MHE. Moreover, this linear force observer cannot separate the dynamic uncertainty (gravitational force) from the estimated torque. Therefore, the operator can feel large feedback force even during free motion that dampens the overall system. The maximum estimated torque in free motion is more than 0.5 Nm but is around 0.4 Nm at the steady state during hard contact, which means that the operator can hardly figure out whether the slave contacts to the environment. The accuracy of position tracking is also influenced by the damped system to a certain extent.

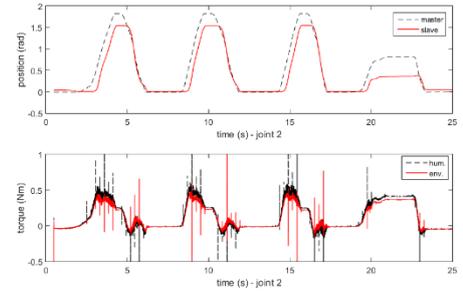

Fig. 2. RFOB (free motion + hard contact)

B. EAOB

The main advantage of the EAOB is that it can efficiently suppress the white measurement noises by using the extended Kalman filter. This observer requires the priori knowledge of the dynamic models. Theoretically, with the ideal dynamic models and ignoring all the colourful noises (dynamic uncertainties), EAOB can accurately estimate the external force provided that the differential of input external force (environmental or human force) is zero. Fig.3 shows the position and torque tracking performance of the system using EAOB. The slave conducts free motion ($0^{th}$ s – $15^{th}$ s) and hard contact ($17^{th}$ s-$23^{rd}$ s). As shown in Fig.3, even applying the relatively accurate dynamic models, the output torque from the EAOB is still not around zero (Maximum 0.4 Nm). It means that the performance of the EAOB is limited by the non-ideal dynamic model. When contacting to the solid wall, the EAOB does not perform very well since the contact force at the transient state is time-varying and it takes a finite settling time for the extended Kalman filter estimate to converge close to the true state. Therefore, the estimated torque varies for seconds until it reaches the steady state. Moreover, requiring calculating the large nonlinear covariance matrix, the EAOB based on three-joint master-slave teleoperation, significantly enlarges the system complexity and calculation. By applying the EAOB for just 3 minutes, the memory of the computer (32 GB) sharply increases from 11% to 70% while the other algorithms (the proposed force observer, RFOB, and NDOB) basically do not occupy any memory. Due to the two reasons, large calculations and gravity-related force output, [19] only provides the experimental results of 2-DOF Phantom devises by ignoring Joint 2.

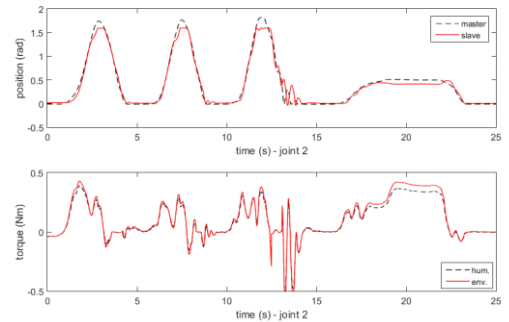

Fig. 3. EAOB (free motion + hard contact)

C. NDOB

The NDOB is another force estimation algorithm that requires accurate dynamic model and the measurement of the velocity. In this experiment, the measurement noises of the velocity are eliminated by the proposed MHE algorithm in order to make a better comparison between NDOB and the proposed force observer. Fig.4 shows the position and torque tracking performance of the system using NDOB. The slave conducts free motion ($0^{th}$ s – $20^{th}$ s) and hard contact ($23^{th}$ s-$27^{th}$ s). Compared with the RFOB, it does not produce large noise (requiring clean velocity measurement without measurement noise); and comparing with the EAOB, it possesses fast state convergence at the transient state. However, requiring ideal dynamic model is still its major shortage. Using the non-ideal dynamic model, large force feedback is still provided by the NDOB during free motion (maximum 0.4 Nm).

From Figs.2-4, one can conclude that the common shortage of the above three force observer is the large force output during free motion that can significantly influence the operator's perception on the environment.

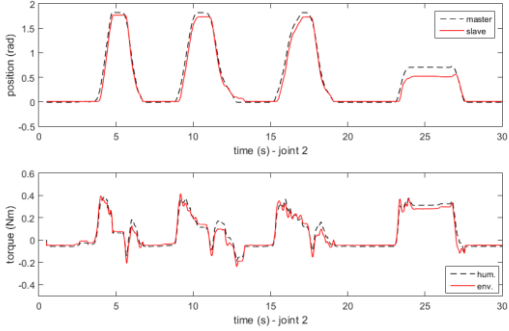

Fig. 4. NDOB (free motion + hard contact)

D. The proposed force observer

Fig.5 shows the position and torque tracking performance of the system using the proposed force observer. The slave conducts free motion ($0^{th}$ s – $18^{th}$ s) and hard contact ($22^{rd}$ s-$28^{th}$ s). The Type-2 fuzzy logic algorithm is applied to model the overall system dynamics and to compensate for the uncertainty. During free motion, the estimated environmental torque is neighbouring zero so that the human applied torque is small (maximum 0.1Nm), which means that basically no force is fed back to the master side and the operator does not feel a damped system. When contacting to the solid wall, the estimated environmental torque fastly increases to 0.4 Nm without large variation possessed by EAOB. Also, the human applied torque also closely tracks the environmental torque during hard contact. Due to the robustness of the Type-2 fuzzy logic strategy, the system dynamic uncertainties (colourful noises) can be efficiently compensated for to the extent that both accurate position and torque tracking is achieved according to Fig.5. Figs.6 and 7 show the position and the velocity output from the position encoder and the velocity estimation algorithm in (13) (measured position and velocity), and the estimated positon and velocity from the proposed MHE algorithm. One can see that the estimated positon signals are exactly the same as the measured position signals in both master and slave. Moreover, the measurement noise (white noise) in the velocity is also largely suppressed.

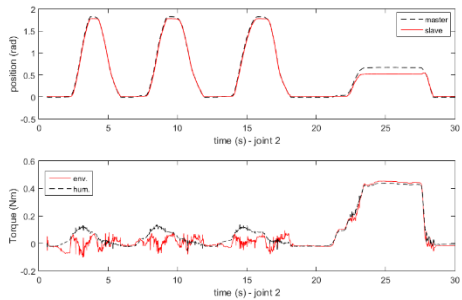

Fig. 5. The proposed force observer (free motion + hard contact)

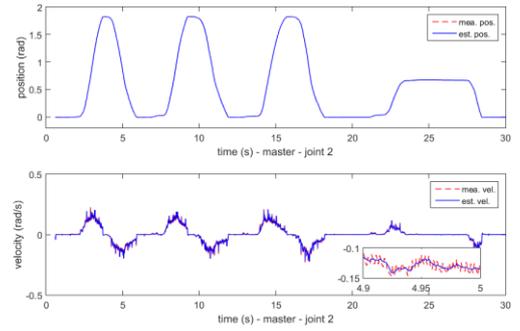

Fig. 6. Measured and estimated position and velocity (Master, free motion + hard contact)

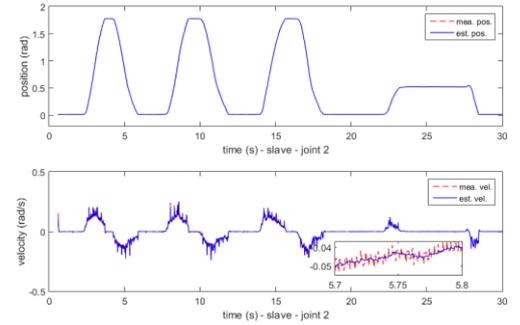

Fig. 7. Measured and estimated position and velocity (Slave, free motion + hard contact)

E. Removing MHE

The measurement noise actually can seriously influence the system stability and performance. Fig. 8 shows the position and velocity of the system using velocity estimation in (13) without MHE and Fig. 9 shows the position and velocity of the system that directly uses differential of the position signals to estimate velocity. One can see that large perturbations are in Figs. 8 and 9. Comparing Figs. 5-7 to Figs. 8-9, the superiority of the proposed MHE is demonstrated.

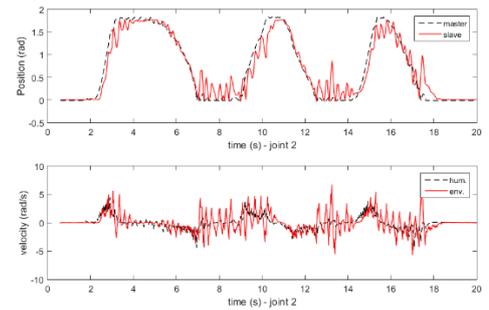

Fig. 8. Teleoperation using velocity estimation in (13) without MHE (free motion)

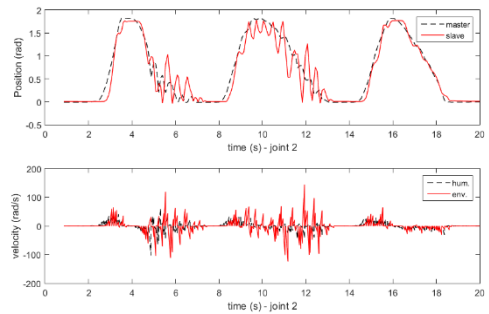

Fig. 9. Teleoperation directly using differential of position signals (free motion)

## 5.2 free motion + soft contact + hard contact

In this subsection, the slave is firstly controlled to conduct free motion, then it contacts to a soft sponge and finally contact to a solid object in Fig.1.

### A. RFOB

Fig. 10 shows the position and torque tracking performance of the system using RFOB. From Fig.10, one can see that the torque variation during free motion, during soft contact and during hard contact is basically the same. Without vision feedback, the operator can hardly distinguish the three scenarios. The output noises are also very large in the all the three scenarios.

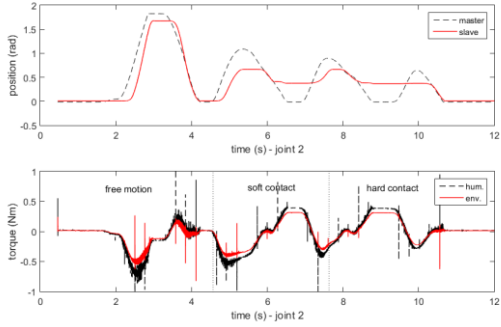

Fig. 10. RFOB (free motion + soft contact + hard contact)

### B. EAOB

In this experiment, the centrifugal force model $C(q,\dot{q})$ and the gravity model $g(\dot{q})$ are removed and the Mass model $M(q)$ is multiplied by 2. Fig.11 shows the position and torque tracking performance of the system using EAOB. From Fig.11, the torque feedback in the free motion is largely enhanced because of the increased velocity while during the soft and the hard contact, there basically no torque output (maximum 0.05 Nm). This experiment shows how large the dynamic model influences the model-based algorithm like EAOB and NDOB. In some complex robotic application where the dynamic model is hardly derived, such these algorithms will not have superiority.

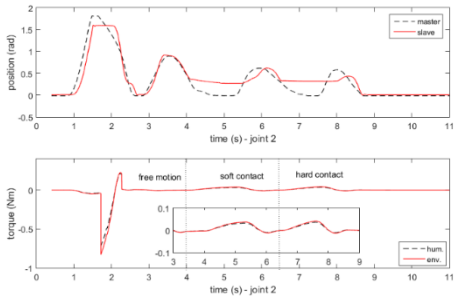

Fig. 11. RFOB (free motion + soft contact + hard contact)

### C. NDOB

In this experiment, only the gravity model $g(\dot{q})$ is removed in the NDOB. Fig.12 shows the position and torque tracking performance of the system using NDOB. During the transient state of the soft contact where velocity is gradually converging to zero, NDOB cannot provide an accurate torque tracking between the master and the slave. Moreover, since the inverse of the nonlinear mass model is required to achieve in the nonlinear dynamic model based observer like NDOB and EAOB, without an ideal mass model, $M(q)$ is probably close to zero in a certain time so that the singularity problem will be produced. This is the reason that sharp signal perturbation occurs in the 4.2$^{th}$ second.

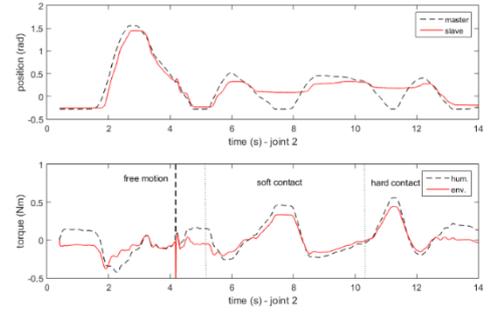

Fig. 12. NDOB (free motion + soft contact + hard contact)

### D. The proposed strategy

Fig.13 shows the position and the torque tracking performance of the system using the proposed force observer. During free motion, the estimated environmental torque is still neighboring zero that allows the operator not to provide large force to drive the slave (Maximum 0.1 Nm). During the soft contact, the estimated environment torque gradually reach 0.3 Nm and during this period, the human torque keeps closely tracking the environment torque. Then during the hard contact, both the environmental and human torque fastly reach 0.4 Nm. This experiment provides the apparently different performances of the proposed force observer during the three scenarios, free motion, soft contact and the hard contact.

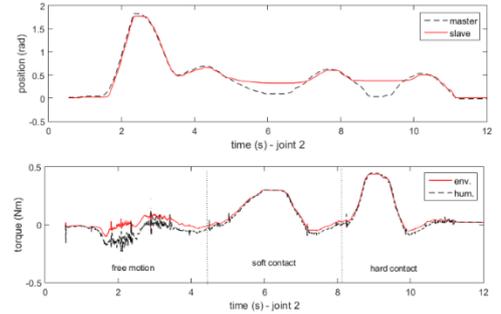

Fig. 13. The proposed force observer (free motion + soft contact + hard contact)

## 6. Conclusion

In this paper, a novel force estimation strategy based on Type-2 fuzzy-logic modelling and the MHE is proposed and is applied to a four-channel force-reflection teleoperation control algorithm. Compared with existing methods, the proposed force observer does not rely on any accurate mathematical functions of the robotic dynamics and is robust to compensate for the colourful noises (dynamic uncertainties) and white noises (measurement noises). By applying the proposed algorithm, the force-reflection transparency of the bilateral teleoperation can be largely improved. The experimental results prove the superiority of the proposed strategy through the comparison with other force estimation algorithms.

## Appendix

*Proof (Theorem 1)*: Set $z_v = \int_0^t (q(\iota) - v(\iota))d\iota$ and define the Lyapunov function $V$ to be

$$V = (\Delta e_v)^T \Delta e_v + k_{v1}k_{v2}z_v^T z_v \quad (A1)$$

Applying (13), the deferential of $V$ can be written as

$$\dot{V} = 2(\Delta e_v)^T(\dot{q} - k_{v1}\Delta e_v - k_{v2}\Delta e_v - k_{v1}k_{v2}z_v) + 2k_{v1}k_{v2}(\Delta e_v)^T z_v$$
$$= -2k_{v1}(\Delta e_v)^T\Delta e_v - 2k_{v2}(\Delta e_v)^T\Delta e_v + 2(\Delta e_v)^T\dot{q}$$
$$\leq -(k_{v1} + k_{v2})(\Delta e_v)^T\Delta e_v + \frac{\dot{q}^T\dot{q}}{k_{v1}+k_{v2}} \quad (A2)$$

From (A2), under the static condition, $\dot{V}$ is definitely non-positive. Under the moving condition, by setting $k_{v1} \gg 1$, $\dot{V}$ can also be guaranteed to be negative semi-definite. Therefore, we can conclude that $\Delta e_v$ belongs to $L_2$ space. *Q.E.D.*

*Proof (Theorem 3):* Consider the following Lyapunov function:
$$V_{e_X}(k) = e_X(k-N)^T P_X e_X(k-N) \quad (A3)$$

which satisfies the condition i) of Definition 1 by choosing $\alpha_1(r) = \lambda_{P_X}^{min} \cdot r^2$ and $\alpha_2(r) = \lambda_{P_X}^{max} \cdot r^2$, where $\lambda_{P_X}^{min}$ and $\lambda_{P_X}^{max}$ are minimum and maximum eigenvalues of $P_X$ respectively. From (29), one can have:
$$V_{e_X}(k) - V_{e_X}(k-1) = e_X(k-N)^T P_X e_X(k-N) - e_X(k-N-1)^T P_X e_X(k-N-1)$$
$$= (\Psi_k e_X(k-N-1) + \xi_k)^T P_X (\Psi_k e_X(k-N-1) + \xi_k) - e_X(k-N-1)^T P_X e_X(k-N-1)$$
$$= e_X(k-N-1)^T (\Psi_k^T P_X \Psi_k - P_X) e_X(k-N-1) + 2e_X(k-N-1)^T \Psi_k^T P_X \xi_k + \xi_k^T P_X \xi_k$$
$$\leq e_X(k-N-1)^T (\Psi_k^T(P_X+I)\Psi_k - P_X)e_X(k-N-1) + \xi_k^T(P_X+P_X^2)\xi_k$$
$$\leq -e_X(k-N-1)^T Q_X e_X(k-N-1) + \xi_k^T(P_X+P_X^2)\xi_k$$
$$\leq -\alpha_3(\|e_X(k-N-1)\|) + \sigma(\|\xi_k\|) \quad (A4)$$

which satisfies the condition ii) of Definition 1 with $\alpha_3(r) = \lambda_{Q_X}^{min} \cdot r^2$ and $\sigma(r) = [\lambda_{P_X}^{max} + (\lambda_{P_X}^{max})^2] \cdot r^2$. Therefore, the error in (29) is *input-to-state stable*. *Q.E.D.*

*Proof (Theorem 4):* Consider the Lyapunov function $V = V_1 + V_2 + V_3 + V_4 + V_5 + V_6$ that
$$V_1 = \frac{1}{2}\hat{q}_m^T M_m \hat{q}_m + \frac{1}{2}\frac{K_m}{K_s}\hat{q}_s^T M_s \hat{q}_s \quad (A5)$$
$$V_2 = \frac{1}{2}K_m(\hat{q}_m(t) - \hat{q}_s(t))^T(\hat{q}_m(t) - \hat{q}_s(t)) \quad (A6)$$
$$V_3 = \int_{-T_1^{max}}^0 \int_{t-\gamma}^t \hat{q}_m^T(\delta)\hat{q}_m(\delta) d\delta d\gamma + \int_{-T_2^{max}}^0 \int_{t-\gamma}^t \hat{q}_s^T(\delta)\hat{q}_s(\delta) d\delta d\gamma \quad (A7)$$
$$V_4 = \frac{1}{2}\left(\tau_{h1}^*(t) - \tau_{e1}^*(t-T_2(t))\right)^T \left(\tau_{h1}^*(t) - \tau_{e1}^*(t-T_2(t))\right) \quad (A8)$$
$$V_5 = \frac{1}{2}\left(\tau_{h3}^*(t-T_1(t)) - \tau_{e3}^*(t)\right)^T \left(\tau_{h3}^*(t-T_1(t)) - \tau_{e3}^*(t)\right) \quad (A9)$$
$$V_6 = \frac{1}{2}\Delta\tau_{h2}^T \Delta\tau_{h2} + \frac{1}{2}\Delta\tau_{e2}^T \Delta\tau_{e2} \quad (A10)$$

The differentials of the above Lyapunov functions are derived as
$$\dot{V}_1 = \hat{q}_m^T(t)\left(K_m\left(\hat{q}_s(t-T_2(t)) - \hat{q}_m(t)\right) - B_m\hat{q}_m(t) + K_h\left(\tau_{h1}^*(t) - \tau_{e1}^*(t-T_2(t))\right) + \Delta\tau_{h2}\right) + \frac{K_m}{K_s}\hat{q}_s^T(t)\left(K_s\left(\hat{q}_m(t-T_1(t)) - \hat{q}_s(t)\right) - B_s\hat{q}_s(t) + K_e\left(\tau_{h3}^*(t-T_1(t)) - \tau_{e3}^*(t)\right) + \Delta\tau_{e2}\right) \quad (A11)$$

$$\dot{V}_2 = K_m \hat{q}_m^T(t)\left(\hat{q}_m(t) - \hat{q}_s(t-T_2(t))\right) + K_m\hat{q}_s^T(t)\left(\hat{q}_m(t-T_1(t)) - \hat{q}_s(t)\right) - K_m\hat{q}_m^T(t)\int_{t-T_2(t)}^t \hat{q}_s(\delta)d\delta - K_m\hat{q}_s^T(t)\int_{t-T_1(t)}^t \hat{q}_m(\delta)d\delta \quad (A12)$$

$$\dot{V}_3 \leq T_1^{max}\hat{q}_m^T(t)\hat{q}_m(t) - \int_{t-T_1(t)}^t \hat{q}_m^T(\delta)\hat{q}_m(\delta)d\delta + T_2^{max}\hat{q}_s^T(t)\hat{q}_s(t) - \int_{t-T_2(t)}^t \hat{q}_s^T(\delta)\hat{q}_s(\delta)d\delta \quad (A13)$$

$$\dot{V}_4 = \left(Y_h\Delta\tau_{h1}(t) + \aleph_{h1}\hat{q}_m(t) - (1-\dot{T}_2(t))Y_e\Delta\tau_{e1}(t-T_2(t)) - (1-\dot{T}_2(t))\aleph_{e1}\hat{q}_s(t-T_2(t))\right)\left(\tau_{h1}^*(t) - \tau_{e1}^*(t-T_2(t))\right) \quad (A14)$$

$$\dot{V}_5 = \left((1-\dot{T}_1(t))Y_h\Delta\tau_{h3}(t-T_1(t)) + (1-\dot{T}_1(t))\aleph_{h3}\hat{q}_m(t-T_1(t)) - Y_e\Delta\tau_{e3}(t) - \aleph_{e3}\hat{q}_s(t)\right)\left(\tau_{h1}^*(t-T_1(t)) - \tau_{e1}^*(t)\right) \quad (A15)$$

$$\dot{V}_6 = \left(\dot{\tau}_h - Y_h\Delta\tau_{h2}(t) - \aleph_{h2}\hat{q}_m(t)\right)\Delta\tau_{h2} + \left(\dot{\tau}_e - Y_e\Delta\tau_{e2}(t) - \aleph_{e2}\hat{q}_s(t)\right)\Delta\tau_{e2} \quad (A16)$$

Using the following inequalities from Lemma 1 in [21]
$$-2\dot{q}_m^T(t)\int_{t-T_2(t)}^t \dot{q}_s(\eta)d\eta - \int_{t-T_2(t)}^t \dot{q}_s^T(\eta)\dot{q}_s(\eta)d\eta \leq T_2^{max}\dot{q}_m^T(t)\dot{q}_m(t) \quad (A17)$$

$$-2\dot{q}_s^T(t)\int_{t-T_1(t)}^t \dot{q}_m(\eta)d\eta - \int_{t-T_1(t)}^t \dot{q}_m^T(\eta)\dot{q}_m(\eta)d\eta \leq T_1^{max}\dot{q}_s^T(t)\dot{q}_s(t) \quad (A18)$$

we derive
$$\dot{V} \leq -\hat{q}_m^T(t)(B_m - (T_1^{max}+T_2^{max})I)\hat{q}_m(t) - \hat{q}_s^T(t)(B_s - (T_1^{max}+T_2^{max})I)\hat{q}_s(t) - \Delta\tau_{h2}^T(t)Y_h\Delta\tau_{h2}(t) - \Delta\tau_{e2}^T(t)Y_e\Delta\tau_{e2}(t) + \hat{q}_m^T(t)(I-\aleph_{h2})\Delta\tau_{h2}(t) + \hat{q}_s^T(t)(I-\aleph_{e2})\Delta\tau_{e2}(t) + (\aleph_{h1}+K_h)\hat{q}_m(t)\left(\tau_{h1}^*(t) - \tau_{e1}^*(t-T_2(t))\right) + \left(\aleph_{e3} + \frac{K_m K_e}{K_s}\right)\hat{q}_s^T(t)\left(\tau_{h3}^*(t-T_1(t)) - \tau_{e3}^*(t)\right) + (1-\dot{T}_1(t))\aleph_{h3}\hat{q}_m(t-T_1(t))\left(\tau_{h1}^*(t-T_1(t)) - \tau_{e1}^*(t)\right) - (1-\dot{T}_2(t))\aleph_{e1}\hat{q}_s(t-T_2(t))\left(\tau_{h1}^*(t) - \tau_{e1}^*(t-T_2(t))\right) + \bar{\xi}_m\Delta\tau_{h2}(t) + \bar{\xi}_s\Delta\tau_{e2}(t) + \left(Y_h\Delta\tau_{h1}(t) - (1-\dot{T}_2(t))Y_e\Delta\tau_{e1}(t-T_2(t))\right)\left(\tau_{h1}^*(t) - \tau_{e1}^*(t-T_2(t))\right) + \left((1-\dot{T}_1(t))Y_h\Delta\tau_{h3}(t-T_1(t)) - Y_e\Delta\tau_{e3}(t)\right)\left(\tau_{h1}^*(t-T_1(t)) - \tau_{e1}^*(t)\right) \quad (A19)$$

Applying (), $\dot{V}$ can be further simplified as:
$$\dot{V} \leq -\hat{q}_m^T(t)(B_m - (T_1^{max}+T_2^{max})I)\hat{q}_m(t) - \hat{q}_s^T(t)(B_s - (T_1^{max}+T_2^{max})I)\hat{q}_s(t) - \Delta\tau_{h2}^T(t)Y_h\Delta\tau_{h2}(t) - \Delta\tau_{e2}^T(t)Y_e\Delta\tau_{e2}(t) + \bar{\xi}_m\Delta\tau_{h2}(t) + \bar{\xi}_s\Delta\tau_{e2}(t) + \left(Y_h\Delta\tau_{h1}(t) - (1-\dot{T}_2(t))Y_e\Delta\tau_{e1}(t-T_2(t))\right)\left(\tau_{h1}^*(t) - \tau_{e1}^*(t-T_2(t))\right) + \left((1-\dot{T}_1(t))Y_h\Delta\tau_{h3}(t-T_1(t)) - Y_e\Delta\tau_{e3}(t)\right)\left(\tau_{h1}^*(t-T_1(t)) - \tau_{e1}^*(t)\right) \quad (A20)$$

Under the condition that the estimated torques can track the real human and environmental torques, $\Delta\tau_{h1-3}$ and $\Delta\tau_{e1-3}$ are neighbouring zero. Therefore, the term $\bar{\xi}_m\Delta\tau_{h2}(t) + \bar{\xi}_s\Delta\tau_{e2}(t) + \left(\Upsilon_h\Delta\tau_{h1}(t) - \left(1-\dot{T}_2(t)\right)\Upsilon_e\Delta\tau_{e1}\left(t-T_2(t)\right)\right)\left(\tau_{h1}^*(t) - \tau_{e1}^*(t-T_2(t))\right) + \left(\left(1-\dot{T}_1(t)\right)\Upsilon_h\Delta\tau_{h3}\left(t-T_1(t)\right) - \Upsilon_e\Delta\tau_{e3}(t)\right)\left(\tau_{h1}^*(t-T_1(t)) - \tau_{e1}^*(t)\right)$ is also neighbouring zero. $\dot{V}$ becomes negative semi-definite with the position tracking error $\hat{q}_m(t) - \hat{q}_s(t)$, the torque tracking errors $\tau_{h1}^*(t) - \tau_{e1}^*(t-T_2(t))$ and $\tau_{h3}^*(t-T_1(t)) - \tau_{e3}^*(t)$ belonging to $L_2$ space. The system is stable and the position and torque tracking errors converge to zero at the steady state. Q.E.D.